\pdfoutput=1
\documentclass[11pt]{article}
\usepackage{booktabs}
\usepackage{graphicx} 
\usepackage{algorithm}
\usepackage{algpseudocode}
\usepackage{acl}
\usepackage{multirow}
\usepackage[inline]{enumitem}
\usepackage{enumerate}
\usepackage{amsfonts}
\usepackage{latexsym}
\usepackage{xspace}
\usepackage{amsmath}
\usepackage[T1]{fontenc}
\usepackage{pifont}
\usepackage{amssymb}
\usepackage{newfloat}
\usepackage{listings}
\usepackage{multirow}
\usepackage{booktabs}
\usepackage{times}
\usepackage[inline]{enumitem}
\usepackage{latexsym}
\usepackage[T1]{fontenc}    
\usepackage[utf8]{inputenc}
\usepackage[inline]{enumitem}
\usepackage[dvipsnames]{colortbl}
\usepackage[utf8]{inputenc}
\usepackage[T1]{fontenc}
\usepackage{babel} 
\usepackage[toc,page]{appendix}
\usepackage{makecell}
\usepackage{soul}
\usepackage{verbatim}
\usepackage{arydshln}
\usepackage{amssymb}
\usepackage{graphicx} 
\usepackage{tcolorbox}
\usepackage[utf8]{inputenc}
\usepackage{adjustbox}
\usepackage{microtype}
\usepackage{inconsolata}

\title{Generate-then-Ground in Retrieval-Augmented Generation \\ for Multi-hop Question Answering}

\definecolor{lemon}{HTML}{FDFFCC}

\definecolor{Gainsboro}{rgb}{0.86, 0.86, 0.86}
\newcommand{\ie}{\emph{i.e.,}\xspace}
\newcommand{\eg}{\emph{e.g.,}\xspace}
\definecolor{Gray}{gray}{0.95}
\definecolor{LightCyan}{rgb}{0.88,1,1}
\newcommand{\code}[1]{{\ttfamily#1}}
\newcommand{\ours}{GenGround\xspace}

\author{
Zhengliang Shi$^1$~~Weiwei Sun$^1$~~Shen Gao$^2$\\ \textbf{Pengjie Ren$^1$~~Zhumin Chen$^1$~~Zhaochun Ren\textsuperscript{$3$}\thanks{$^*$ Corresponding author.}} \\
$^1$Shandong University, Qingdao, China \\  
$^2$University of Electronic Science and Technology of China, Chengdu, China\\
$^3$Leiden University, Leiden, The Netherlands \\
\texttt{shizhl@mail.sdu.edu.cn~~z.ren@liacs.leidenuniv.nl}\\
}

\begin{document}
\maketitle
\begin{abstract}
Multi-Hop Question Answering (MHQA) tasks present a significant challenge for large language models (LLMs) due to the intensive knowledge required. 
Current solutions, like Retrieval-Augmented Generation, typically \textit{retrieve} potential documents from an external corpus to \textit{read} an answer.
However, the performance of this \textit{retrieve-then-read} paradigm is constrained by the retriever and the inevitable noise in the retrieved documents.
To mitigate these challenges, we introduce a novel \textit{generate-then-ground} (\ours) framework, synergizing the parametric knowledge of LLMs and external documents to solve a multi-hop question.
\ours empowers LLMs to alternate two phases until the final answer is derived:
(1) formulate a simpler, single-hop question and directly generate the answer;
(2) ground the question-answer pair in retrieved documents, amending any wrong predictions in the answer.
We also propose an instructional grounding distillation method to generalize our method into smaller models.
Extensive experiments conducted on four datasets illustrate the superiority of our method.

\end{abstract}

\section{Introduction}

\begin{figure}[t]
        \centering
\includegraphics[width=0.5\textwidth]{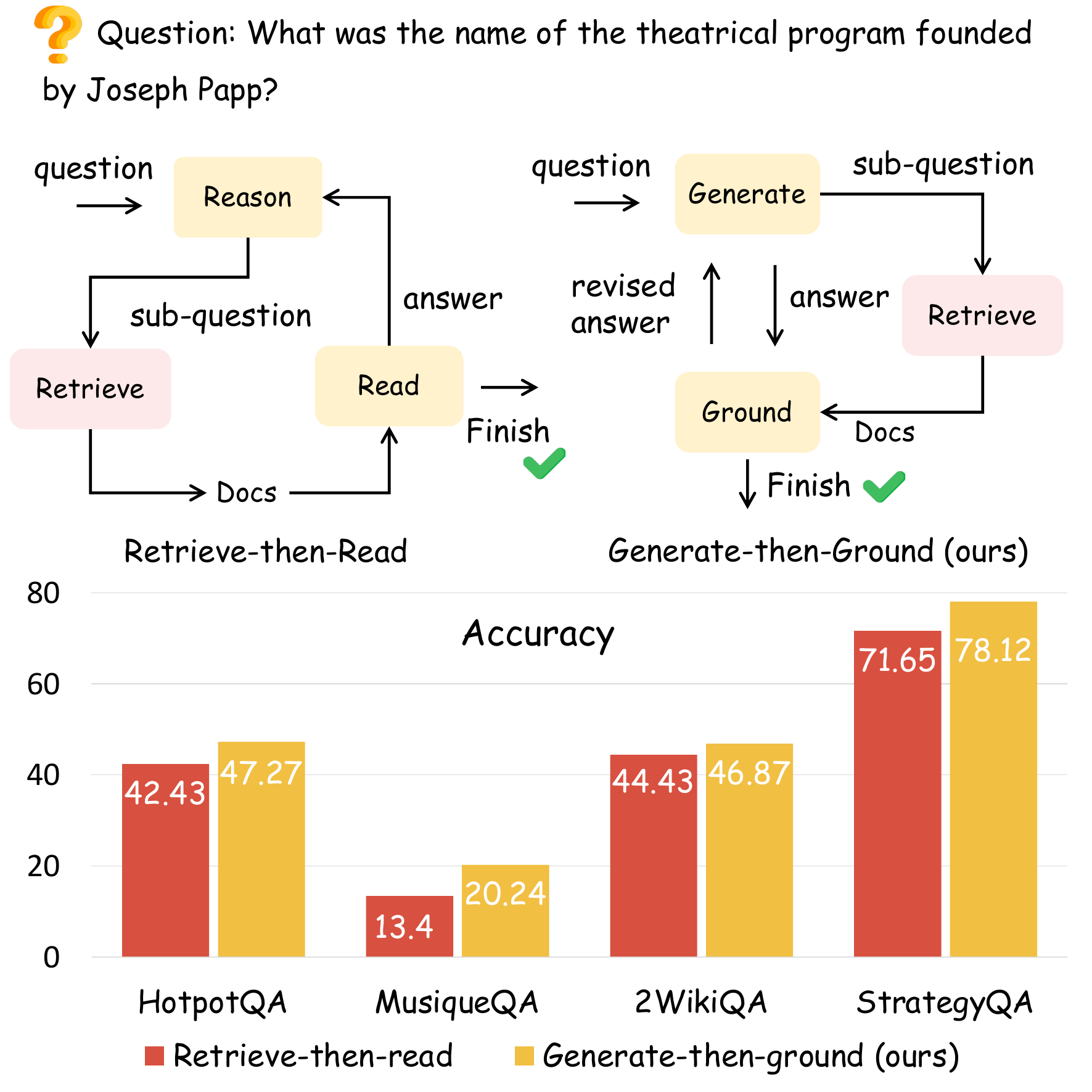}
        \caption{The top block depicts the comparison with the commonly-used \textit{retrieve-then-read} paradigm in MHQA task. The bottom block provides the performance of our method and baselines in four MHQA benchmarks.
        }
\label{fig:intro}
\end{figure}

Multi-Hop Question Answering (MHQA) tasks~\citep{yang2018hotpotqa} require multi-hop reasoning using intensive knowledge to derive the answer~\citep{xu2023searchinthechain}. It has been widely employed in various practical scenarios and domains~\citep{mavi2022survey}.
To answer a multi-hop question, most prior work integrates large language models (LLMs) with information retrieval techniques, following a \textit{retrieve-then-read} paradigm~\cite{Shao2023EnhancingRL}.
As illustrated in Figure~\ref{fig:intro}, the initial step employs the LLMs to break down the complex question and formulate a series of simpler, single questions in a step-by-step manner.
For each step, the LLMs are guided to derive an answer from the relevant documents, which are retrieved using the formulated question~\citep{yao2023react}.

Despite the progress of the \textit{retrieve-then-read paradigm}, it faces \textit{two challenges} in practice. 
First, its effectiveness is constrained by the performance of the retriever~\citep{yu2023generate,xu2023searchinthechain}.
Given that the answer can only be derived from the retrieved documents, the inherent world knowledge of LLMs is overlooked.
This limitation is particularly magnified in multi-hop QA tasks that frequently require complex logical reasoning~\citep{mavi2022survey}. The retrievers may struggle to retrieve all necessary documents to answer the question, leading to a performance decline using this paradigm~\citep{yu2023generate,abdallah2023generator}.
Second, the retrieved documents inevitably contain irrelevant or plausible-looking statements~\citep{,gao2023retrieval,jiang2023active}.
Directly incorporating them into the chain-of-reasoning of LLMs may mislead the LLMs to generate incorrect or irrelevant responses~\citep{adlakha2023evaluating,Thakur2023NoMIRACLKW}.
Therefore, developing an adaptive method for utilizing the retrieved documents remains an active research area.

Inspired by the extensive knowledge and powerful deduction capability of LLMs, \eg ChatGPT or GPT-4, we propose to address these challenges with a novel \textit{generate-then-ground} (\ours) method, as shown in Figure~\ref{fig:intro}.
This approach diverges from the retrieve-then-read paradigm by first allowing the LLMs to generate an immediate answer, and then grounding this answer in evidence to revise it.
In this work, we focus on the following research questions:
(1) \textbf{RQ1}: How does our method synergize the world knowledge of LLMs and retrieved documents to answer a multi-hop question?
(2) \textbf{RQ2}: For LLMs with different scales of parameters, how do we generalize our method?

To address \textbf{RQ1}, we enable LLMs to alternate between \textit{answer deduction} and \textit{instructional knowledge grounding}. In the deduction phase, LLMs form sub-questions from the input question and context. The LLMs then produce an immediate answer for each sub-question. To prevent non-factual hallucination by LLMs~\citep{zhang2023siren,gao2023rarr}, we guide LLMs to revise the answer in the grounding phase, using documents retrieved from an external corpus like Wikipedia. The LLMs ground the question-answer pair in evidence by citing relevant content and correcting errors. This revised answer is used for the next iteration's sub-question, continuing till the final hop. We also introduce a batch grounding strategy for efficient document use.

To address \textbf{RQ2}, we propose an instructional grounding distillation method. Despite LLMs like ChatGPT performing well with our method, smaller models may struggle with instruction-following in the grounding phase. Thus, we use 50k single-hop questions from the Natural Questions (NQ) dataset~\citep{kwiatkowski2019natural}, each with ground-truth and noise documents. We guide ChatGPT to generate and adjust an answer for each question, then distill ChatGPT's process into a student model using instruction tuning.

Extensive experiments on four commonly-used MHQA datasets demonstrate superior performance over strong baselines (\eg ReAct and DSPy), achieving the best performance overall.
We also observe that our instructional grounding distillation empowers the smaller model with strong performance.

To sum up, our main contributions are as follows:
 (1) We propose a novel generate-then-ground framework for retrieval-argument generation technique in multi-hop question tasks, which effectively synergizes the knowledge of LLMs and retrieved documents.
 (2) We introduce an instructional grounding distillation method, enabling a smaller model with the generate-then-ground framework. 
(3) Experiments on four datasets are conducted to demonstrate the superiority of our method.

\section{Related work}

\subsection{Multi-hop Question Answering}

Multi-Hop Question Answering (MHQA) tasks focus on answering questions that require gathering information from multiple sources and conducting multi-step reasoning to arrive at a comprehensive answer~\citep{zhang2023beam,li2023leveraging}.
Some works utilize the knowledge deduction capability~\citep{Wei2022ChainOT} of LLMs to decompose the input question into single-hop questions and then solve them step-by-step~\cite {wang2022self,wang2023query2doc}.
And many techniques~\citep{yao2023tree,besta2023graph} are proposed to improve the reasoning ability of LLMs.
However, the LLMs suffer from generating non-factual statements.
As an intuitive solution, many recent works integrate retrieval into the chain of thought reasoning process~\citep{yao2023react,schick2023toolformer}, prompting the LLMs to generate the answer using retrieved documents.
Although promising, the inevitable noise in retrieved documents could mislead the  LLMs to a wrong reasoning direction and derive a wrong answer~\citep{xu2023searchinthechain}.
In this work, we propose an instructional knowledge grounding method that enables LLMs to find the most relevant evidence from the document list, thereby reducing the effect of the noise.

\subsection{Retrieval-Augmented Generation}
Retrieval-augmented generation (RAG) has been proven a promising technique to improve the performance of LLMs in knowledge-intensive NLP tasks~\citep{zhu2023large,yu2023chain}, which enhances LLMs with retrievers to access external knowledge.
Existing RAG methods typically follow a retrieval-then-read pipeline~\citep{ma2023query,feng2023retrieval,gao2023retrieval}.
Given a query, a retriever is first employed to retrieve the relevant document and a reader is then used to predict the answer on the condition of retrieved documents~\citep{khattab2023dspy}.
Despite the advancement of this paradigm, it is limited by the accuracy of retrievers~\citep{yu2023generate,zhang2023beam}.

Recent works mitigate this problem by incorporating the world knowledge of LLMs~\citep{gao2023rarr,chen2023beyond}, where they utilize the LLM as a knowledge base to generate contextual documents and then read the answer from both retrieved and generated documents~\citep{abdallah2023generator}.
However, the potential knowledge conflict between the two knowledge sources is ignored, which may hallucinate the LLMs~\citep{xie2023adaptive,thakur2023nomiracl,mallen2023not}.
In our work, we propose a generate-then-ground framework, more effectively incorporating the parametric knowledge of LLMs and external documents.

\section{Generate-then-Ground with LLMs}\label{sec:method-gpt}
\begin{figure}[t]
        \centering
	\includegraphics[width=0.495\textwidth]{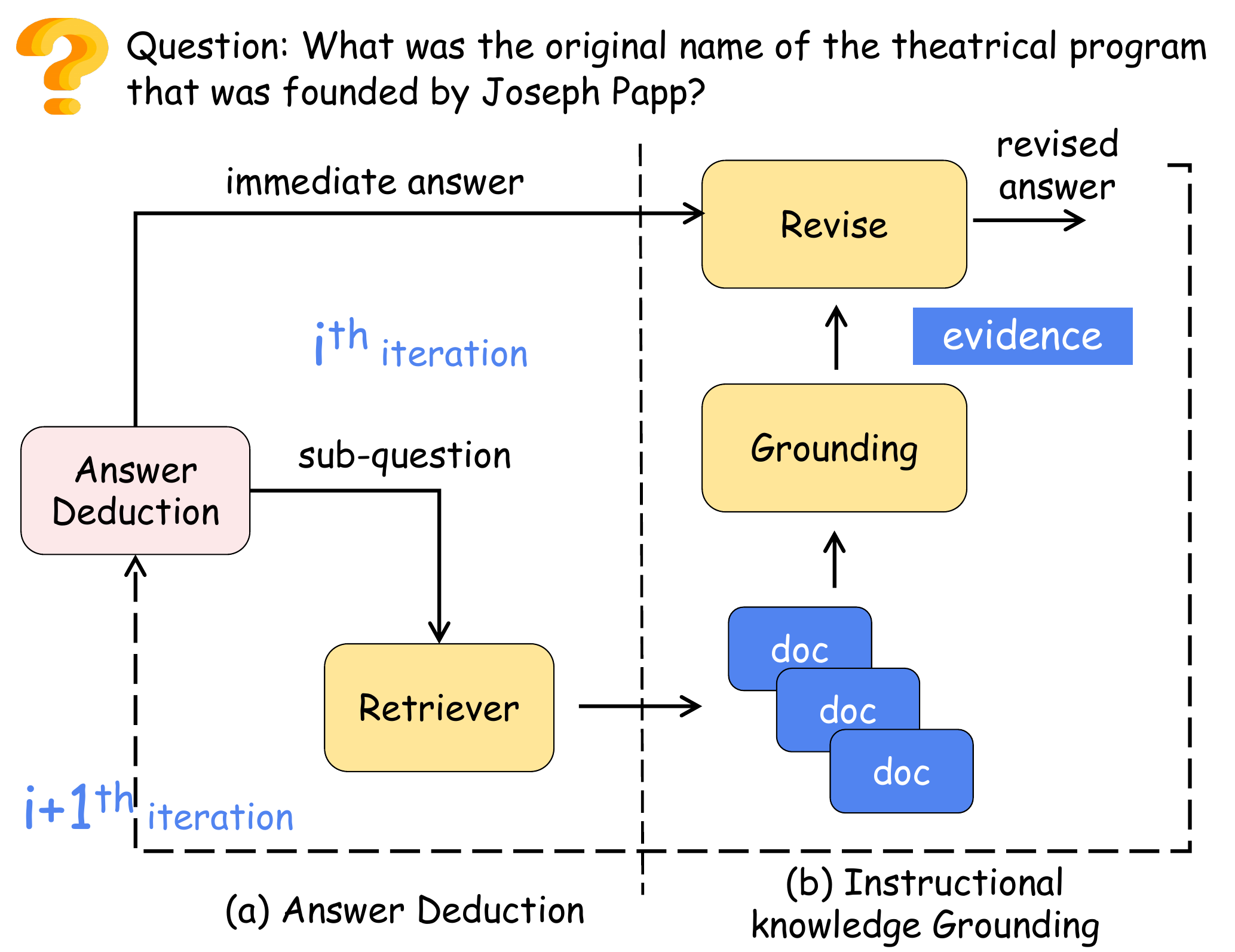}
        \caption{The architecture of the proposed  generate-then-ground framework.}
 \label{fig:method}
\end{figure}

\begin{figure}[t]
        \centering
	\includegraphics[width=0.47\textwidth]{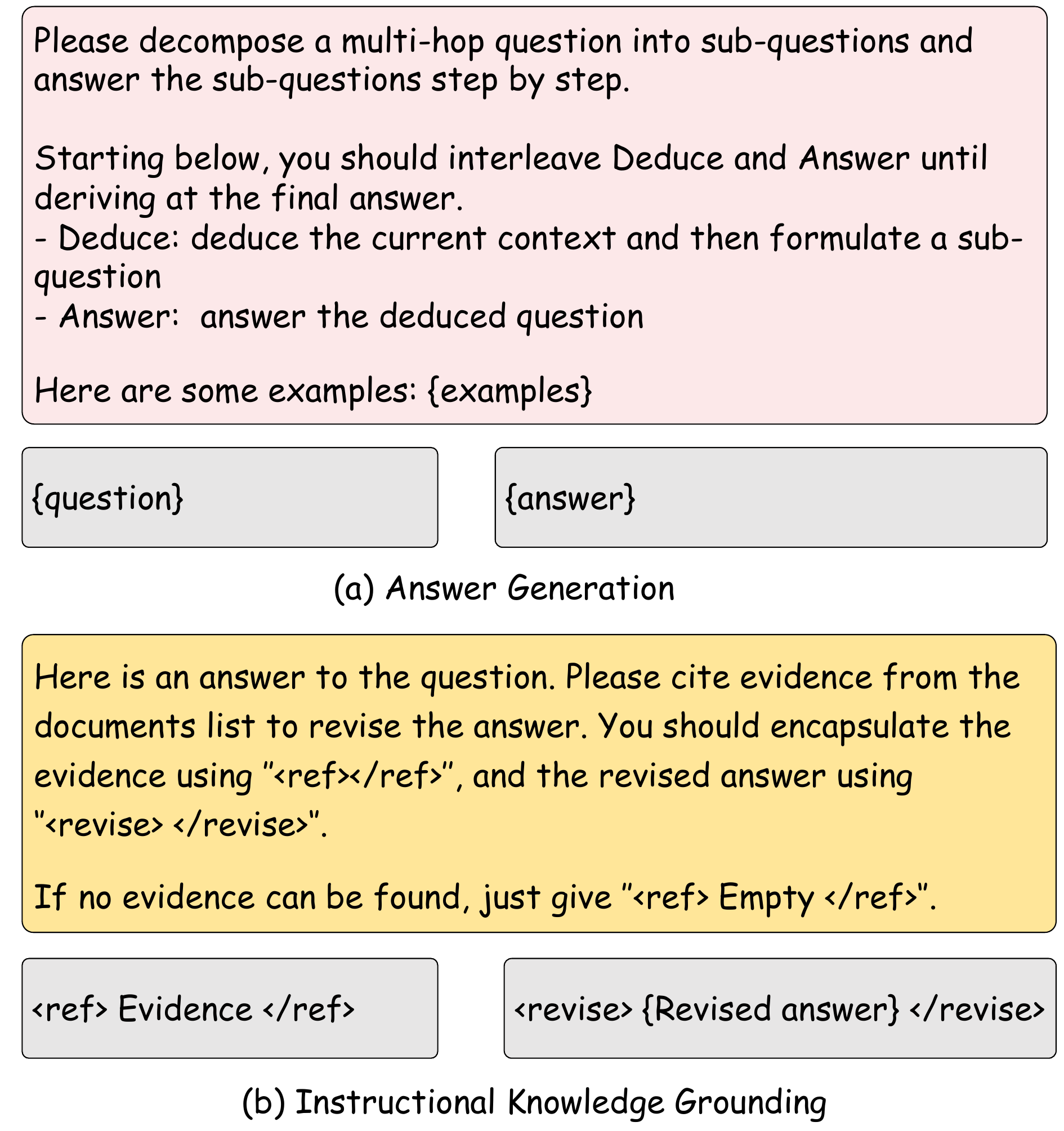}
        \caption{
        The instruction for the \textit{answer deduction }(a) and \textit{instructional knowledge grounding}(b) phases in our framework.
        The pink and yellow blacks indicate the input  while the gray blocks indicate the output.}
 \label{fig:prompt}
\end{figure}

This section provides a detailed explanation of \ours.
As depicted in Figure~\ref{fig:method}, \ours empowers the LLMs to alternate between two phases over multiple iterations until the final answer $a$, is derived. These phases include \textit{answer deduction} (Section~\ref{sec:generation}) and \textit{instructional knowledge grounding} (Section~\ref{sec:grounding}).
During each iteration, the former guides the LLMs to generate a simpler, single-hop answer. The latter phase addresses LLMs' non-factual hallucination~\citep{zhang2023siren} by prompting them to ground the question-answer pair in evidence and correct wrong predictions. The revised answer and sub-question are then integrated into the LLMs' context for the following iteration's prediction.

\subsection{Answer Deduction}\label{sec:generation}

The answer deduction phase aims to utilize the world knowledge of LLMs stored within their parameters $\theta$.
Given the complex reasoning involved in multi-hop questions~\citep{tang-etal-2021-multi,li2023leveraging}, it can be challenging to generate an accurate answer directly.
As a result, we guide the LLM, denoted as $\mathcal{M}_\theta$, to break down a complex question $\mathcal{Q}$ into single-hop questions with a fine granularity.
Formally, for $i$-th iteration,  we denote the current context as $\mathcal{H} = \{(q_j,\Tilde{j}_i) | j < i\}$, which comprises the accumulation of previous deduced sub-questions, $q_{<i}$, and  revised answers, $\Tilde{a}_{<i}$.
The context $\mathcal{H}$, along with the input question $\mathcal{Q}$, is then fed into the LLMs to generate a sub-question, $q_i$, that defines the specific information to be retrieved.
Subsequently, we prompt the LLMs, $\mathcal{M}_\theta$, to directly generate an answer, $a_i$, for the formulated question, $q_i$. This can be formulated as follows:
\begin{equation}
    q_i, a_i = \mathcal{M}_\theta(\mathcal{I_A},\mathcal{Q},\mathcal{H}_i).
    \label{eq:deduce}
\end{equation}
Here, the $\mathcal{I_A}$ represents the instruction shown in Figure~\ref{fig:prompt}, which includes the task demonstration and in-context learning examples.

\begin{figure}[t]
        \centering
	\includegraphics[width=0.47\textwidth]{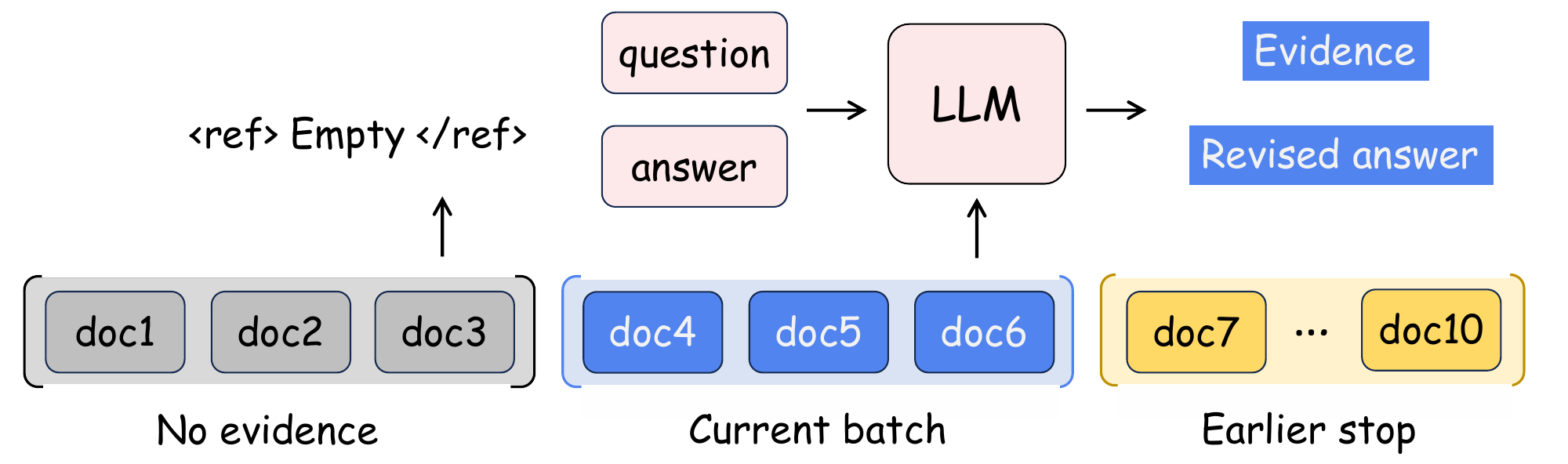}
        \caption{Demonstration of our batch grounding strategy with the batch size of 3 and retrieved documents amount of 10, where the LLMs ground the input question-answer pair into the second batch.}
 \label{fig:batch}
\end{figure}

\subsection{Instructional Knowledge Grounding }\label{sec:grounding}

Given that LLMs may generate non-factual statements or ``hallucinations'', we further guide the LLMs to revise the generated answer, $a_i$, with the support of retrieved documents.
Specifically, in the $i$-th iteration, we initially utilize a retriever (\eg  Google, BM25, or dense retrieval model) to retrieve relevant documents, $\mathcal{D}_i$, using the deduced sub-question $q_i$ (Equation~\ref{eq:deduce}):
\begin{equation}
\mathcal{D}=\text{Retrieval}(q_i).
\end{equation}
We then guide the LLMs, $\mathcal{M}_\theta$, to ground the question-answer pair, $(q_i,a_i)$, in established evidence by citing the most relevant content, $\Tilde{d}$, from the retrieved documents, $\mathcal{D}_i$. Subsequently, we revise the answer $a_i$:
\begin{equation}
       \Tilde{a}_i = \mathcal{M}_\theta(\mathcal{I_G},Q,q_i,a_i).
    \label{eq:ground}
\end{equation}
As depicted in Figure~\ref{fig:prompt}(b), $\mathcal{I_G}$ represents our grounding instruction in a zero-shot setting, while $\Tilde{a}_i $ is the revision trajectory, which includes the cited evidence and the revised answer.
The revision trajectory $\Tilde{a}_i$, along with the question $q_i$, are then combined to build the context, $\mathcal{H}_{i+1}=\mathcal{H}_i \cup \{(q_i,\Tilde{a}_i)\}$, for the LLMs in the subsequent iteration.
If no relevant content can be cited (\eg the citation is \code{Empty}), we keep the generated answer as the revised answer without any changes.

 \subsection{Batch Knowledge Ground}\label{sec:batch}
Since retrieved documents are typically lengthy and contain inevitable noise~\citep{xu2023searchinthechain},
the LLMs are susceptible to being misled by plausible-looking statements during the grounding phase~\citep{sun2023contrastive,thakur2023nomiracl}.
Therefore, we propose a simple yet efficient 
batch grounding strategy.
Suppose the batch size is $b$. We first utilize the LLMs to revise the generated answer $a_i$ using the $(1,b)$-th documents.
If relevant evidence can be cited to revise the answer, we end our grounding phase for the current iteration and move to the next iteration.
Otherwise, we prompt the LLMs to generate an ``Empty'' signal and then access the $(b+1,2b)$-th documents sequentially.
This process continues until the relevant evidence can be found to support our grounding phase. 
Figure~\ref{fig:batch} shows a concrete example with ten retrieved documents.
If no relevant document can be found, we directly output the generated answer as a backup.

\section{Generalization with Grounding Distillation}\label{sec:method-model}
While LLMs like ChatGPT are skilled and adept at following instructions, they are often considered black boxes~\citep{qin2023webcpm,gao2023confucius} and their extensive parameters can increase latency and inference cost in real-world applications~\citep{sun2023chatgpt}. Thus, we aim to adapt our framework to smaller, open source models with fewer parameters. Initial experiments show these smaller models struggle to cite relevant evidence during the knowledge grounding phase. To overcome this, we introduce \textbf{I}nstructional \textbf{G}rounding \textbf{Di}stillation (IDG), which distills the output trajectory of ChatGPT into a smaller student model.

\subsection{Synthesize the Training Dataset}
The instructional grounding distillation collects the trajectory of LLMs, \ie ChatGPT, during the instructional knowledge grounding (Section~\ref{sec:grounding}). This trajectory is then used as the training dataset to distill the grounding capability into a student model.
To achieve this, we first sample 50k questions from the Natural Questions (NQ) dataset~\citep{kwiatkowski2019natural}.
Each question $q$ is paired with a corresponding ground-truth document $\Tilde{d}$ and the noise documents  $\mathcal{D}$.
The questions in the NQ dataset are of high quality and single-hop, making them inherently similar to the setting of our instruction knowledge grounding (Section~\ref{sec:grounding}).
Next, we supplement each question with an immediate answer $a$ and a detailed revision trajectory $\Tilde{a}$.
Specifically, the immediate answer $a$ is generated directly by feeding the question $q$ into a smaller model, such as Mistral-7B~\citep{mistral}.
The revision trajectory $\Tilde{a}$ is generated by ChatGPT with the assistance of the ground truth document.
Various heuristic methods are also used to filter low-quality output (see Appendix \uppercase\expandafter{\romannumeral3} for more details). 
The statistics of out synthetic dataset is provided in Table~\ref{tab:dataset}.

\begin{table}[t]
\centering
\begin{adjustbox}{width=\columnwidth,center}
\begin{tabular}{p{8cm} r}
\toprule
\textbf{Statistic}
\\
\midrule
\# The data scale &  45,710 
\\
\# The average length of input instruction    & 70.87
\\
\# The average length of output   & 683.21
\\
\# The average number of ground truth documents & 1.00
\\
\# The average length of ground truth documents & 117.57
\\
\bottomrule
\end{tabular}
\end{adjustbox}
\caption{The statistics of our synthetic dataset in the instructional grounding distillation method.}
\label{tab:dataset}
\end{table}

\subsection{Training Objective}
Formally, for each question $q$ in our synthetic dataset, we train the model to cite the relevant content from a document list and revise any incorrect predictions in the immediate answer $a$ following the instruction $\mathcal{I_G}$.
Using the collected revision trajectory $\Tilde{a}$, 
we apply the standard language modeling loss to optimize the student model:
\begin{equation}
    \begin{aligned}
       \mathcal{L_G}
       & = - \log P_\theta (\Tilde{a}|\mathcal{I_G}, \{\Tilde{d}\} \cup \mathcal{D})  \\
       & =  - \sum_{t=1}^{|\Tilde{a}|} \log P_\theta(\Tilde{a}_t | \Tilde{a}_{(<t)}, \mathcal{I_G}, \{\Tilde{d}\} \cup \mathcal{D}).
   \end{aligned}
    \label{eq:qe}
\end{equation}
Here, the  $\Tilde{d}$  indicates the ground-truth document, and $\mathcal{D}$ indicates the noise documents.

\begin{table*}[htbp]
\centering
\begin{adjustbox}{width=2.05\columnwidth,center}
\setlength\tabcolsep{4pt}
\begin{tabular}{@{}p{5.2cm} ccc ccc  ccc c@{}}

\toprule
\multirow{3}{*}{\textbf{Methods}} 
& \multicolumn{3}{c}{\textbf{HotpotQA}} 
& \multicolumn{3}{c}{\textbf{MuSiQue}} 
& \multicolumn{3}{c}{\textbf{2Wikimultihopqa}} 
& \multicolumn{1}{c}{\textbf{StrategyQA}} 
\\
\cmidrule(lr){2-4} \cmidrule(lr){5-7} \cmidrule(lr){8-10}  \cmidrule(lr){11-11} 
&  F1
&  Acc
& Acc$\dagger$
&  F1
&  Acc
& Acc$\dagger$
&  F1
&  Acc
& Acc$\dagger$
&  Acc
\\

\hline
\rowcolor{Gainsboro}\multicolumn{11}{l}{\textit{Generate w/o Retrieval}} \\

CoT~\citep{Wei2022ChainOT}
&   35.28 &	30.79	&	37.07
& 23.35 &	13.21	&	17.85
&  35.41 & 32.46	&	34.52
& 67.83	
\\

CoT-SC~\citep{Wang2022SelfConsistencyIC}
& 42.25  &	38.68  &  39.07 
& 15.61 &	10.02	&	12.42
& 40.37 &	36.57 	&	38.59
& 70.84	
\\

GenRead~\citep{yu2023generate}
& 35.21 &	36.81 	&	37.54
& 9.77 &	9.29 		& 10.32
&23.13 &	20.62	&  28.31
&	67.13
\\

GenRead \textit{w/ decomposition}
& 42.28 & 43.32 &  45.31
& 20.13 & 17.58 	&20.62	 
& 41.19 &	41.63	&	43.24
&	68.13
\\

\hline
\rowcolor{Gainsboro}\multicolumn{11}{l}{\textit{Generate w/ Retrieval}} \\

VE~\citep{zhao-etal-2023-verify}
& 29.64  &	22.64 &  24.64 
& 6.5	& 11.14 & 15.57 
&  13.76 	& 31.57 &  32.64
& 	63.07
\\

ReAct~\citep{yao2023react}
& 40.70 & 33.10 & 37.12
&  15.34 & 17.32 &  19.32
& 35.50  &  30.10  & 33.41
& 68.37
\\

GRG \textit{w/ decomposition}
&  50.21 &	45.18 & 50.80	
& 24.87   & 17.91	 & 22.33
&  40.42 &	 40.48	&43.05	 
& 75.21
\\

RetGen~\citep{Shao2023EnhancingRL}
&  28.30 & 41.04 &  44.10 
& 21.04 & 17.69  &  20.19
&  36.00   & 42.17 & 45.21
& 73.42
\\

SearChain~\citep{xu2023searchinthechain}
& - & 46.76	 & 48.12
& - &	17.07  &  20.45
& - &  42.14 & 46.27	
& 	76.95	
\\

DSPy~\citep{khattab2023dspy}
& 47.80 &	42.43	& 50.07	
& 20.11 &	13.40 & 17.40
& 44.77  &	43.43	& 45.43
&	71.78
\\

\midrule
\textbf{GenGround (Ours)} 
&  \textbf{52.26} & \textbf{47.27}  & \textbf{55.73}
&	\textbf{27.36} & \textbf{20.24} &  \textbf{24.77}
& \textbf{50.21} & \textbf{45.61} & \textbf{48.58} 
& \textbf{77.12}
\\

\bottomrule
\end{tabular}
\end{adjustbox}
\caption{
Evaluation results on multi-hop question answering datasets. Acc$\dagger$ indicates the semantic accuracy of model outputs evaluated with \code{gpt-3.5-turbo-instruct} with the same prompt.
Since the \textit{SearChain} prompts the LLM to generate a long-form answer while the ground truth answer in our dataset is short-form,  we only evaluate it with the Acc and Acc$^\dagger$ metrics.
}
\label{tab:main}

\end{table*}
\begin{table}[!t]
\centering
\begin{adjustbox}{width=\columnwidth,center}
\begin{tabular}{p{2.3cm} ccc c}
\toprule
\textbf{Method} 
& \textbf{HQA} 
& \textbf{MQA} 
& \textbf{WQA}
& \textbf{Average $\Delta$} \\

\hline
\rowcolor{Gainsboro}\multicolumn{5}{l}{\textit{Retrieval $\rightarrow$ ColBERTv2}} \\

Ours (IDG)
& 42.08
& 14.37
& 32.69
& -
\\

Ours (Vanilla)
& 38.31
& 11.34
& 29.45
& 3.35$\downarrow$
\\

DSPy
& 36.41
& 7.42
& 28.31
& 5.67$\downarrow$
\\

GRG \textit{w/ dp}
& 32.56
& 9.34
& 25.63
& 7.20$\downarrow$
\\

RetGen
& 26.52
& 10.12
& 24.13
&  9.45$\downarrow$
\\

SearChain
& 24.62 
& 9.46
& 26.53
& 9.51$\downarrow$
\\

\bottomrule
\end{tabular}
\end{adjustbox}
\caption{Accuracy (Acc) on three datasets with Mistral-7B as backbone.
The \textit{w/ dq} indicates  \textit{decomposition}.
The \textit{Vanilla} and \textit{IDG} indicate enable our framework by prompting and our grounding distillation, respectively.
}
\label{tab:open-model}
\end{table}

\begin{table}[!t]
\centering
\begin{adjustbox}{width=\columnwidth,center}
\setlength\tabcolsep{4pt}
\begin{tabular}{@{}p{2.2 cm} ccc c@{}}

\toprule
\multirow{2}{*}{\textbf{Ablation}} 
& \multicolumn{3}{c}{\textbf{HQA}} 
& \multicolumn{1}{c}{\textbf{SQA}} 
\\
\cmidrule(lr){2-4}  \cmidrule(lr){5-5} 
&  F1
&  Acc
& Acc$\dagger$
&  Acc
\\

\midrule

\textit{w/o deduction}
& 42.65 $\downarrow_{9}$ 
&  41.08$\downarrow_{6}$ 
&43.14 $\downarrow_{12}$ 
& 66.51 $\downarrow_{10}$ 
\\

\textit{w/o grounding}
& 45.14$\downarrow_{7}$ 
& 41.35$\downarrow_{4}$ 
& 43.23$\downarrow_{5}$
& 72.34$\downarrow_{5}$
\\

\textit{w/o batch}
& 47.27$\downarrow_{5}$ 
& 45.03$\downarrow_{2}$ 
& 51.19$\downarrow_{4}$ 
& 71.72$\downarrow_{5}$ 
\\

\bottomrule
\end{tabular}
\end{adjustbox}

\caption{Evaluation results of our ablation study on two MHQA benchmarks.}
\label{tab:ablation}
\end{table}

\section{Experimental Setup}\label{sec:experiment}

\subsection{Datasets}
In line with previous research~\citep{Lewis2020RetrievalAugmentedGF,xu2023searchinthechain}, we conduct experiments on four common-used MHQA benchmarks, namely
HotpotQA (HQA)~\citep{yang2018hotpotqa},  MuSiQue (MQA)~\citep{trivedi2021musique}, 2Wikimultihopqa (WQA)~\citep{xanh2020_2wikimultihop}, and 
StrategyQA (SQA)~\citep{geva2021strategyqa}.
StrategyQA is derived from BIG-bench\footnote{\url{https://github.com/google/BIG-bench}}.
For the remaining benchmarks, we randomly sample 1.4k questions adhering to RetGen~\citep{Shao2023EnhancingRL} and Verify-and-Edit~\citep{zhao-etal-2023-verify}.

\subsection{Baselines}
We compare our method with  \textit{Generation w/ Retrieval} and \textit{Generation w/o Retrieval} methods, respectively. The \textit{w/} indicates \textit{with} while the \textit{w/} indicates \textit{without}.

The \textit{Generation w/o Retrieval} methods utilize the parametric knowledge of LLMs to answer questions. This includes
(1) \textit{CoT}~\citep{Wei2022ChainOT}, which prompts the LLMs to a series of intermediate reasoning steps when answering a question;
(2) \textit{CoT-SC}~\citep{Wang2022SelfConsistencyIC}, which samples a diverse set of reasoning paths and then selects the most consistent answer;
and (3) \textit{GenRead}~\citep{yu2023generate}, which generates the answer by reading from the documents generated by LLMs. 
The \textit{Generation w/ Retrieval} methods augment LLMs with retrievers to access external knowledge when answering questions, including:
(1) ReAct~\citep{yao2023react}, which interleaves question generation, document retrieval, and knowledge incorporation to answer a question;
(2) GRG~\citep{abdallah2023generator}, reading an answer from both retrieved documents and contextual documents generated by LLMs;
(3) RetGen~\citep{Shao2023EnhancingRL}, which enhances LLMs with an iterative retrieval-generation synergy strategy to answer a multi-hop question.
(4) DSP~\citep{khattab2022demonstrate}, a programming framework empowered by LLMs; 
and (5) SearChain~\citep{xu2023searchinthechain}, which dynamically interacts with the retriever to verify and correct the generated answers.

Considering the complexity of multi-hop questions, we enhance GenRead and GRG with the chain-of-thought technique (\textit{w/ decomposition}), dividing the input question into sub-questions and using GenRead (or GRG) for each sub-question.

\subsection{Evaluation Metrics}
Following previous studies~\citep{xu2023searchinthechain,ren2023investigating}, we use \textit{Accuracy} (Acc) and \textit{F1} metrics for evaluation. The \textit{accuracy} metric checks if the ground truth answer is in the generated answer, which is also named cover-EM.
The \textit{F1} score is used to measure the overlap between the generated answer and the ground truth answer, which represents the harmonic mean of precision and recall. Recall is determined by considering the number of overlaps with the correct answer tokens, while precision is determined by considering the number of overlaps with all generated tokens.
We also assess semantic accuracy (Acc$^\dagger$) using gpt-3.5-turbo-instruct\footnote{\url{https://openai.com/}\label{openai}} for a more thorough evaluation, which prompts the LLMs to evaluate the correctness of the generated answer taking the ground truth answer as reference.
In this work, we implement the Acc$^\dagger$ using gpt-3.5-turbo-instruct and the full prompt for our evaluation can be found in Appendix \uppercase\expandafter{\romannumeral1}.

To counter the potential bias of automatic metrics~\citep{Shi2023RADERD}, we conduct a human evaluation, with three educated individuals assessing the \textit{correctness} of 120 randomly sampled cases from four benchmarks on a three-scale rating.

\subsection{Implementation Details}
We utilize OpenAI's \textit{gpt-3.5-turbo} as the backbone for our framework and all baselines, with the decoding temperature set to 0 for deterministic generation, and batch size set to 3 in our batch grounding strategy. The open source model, Mistral-7B~\citep{mistral}, is also employed for comparison. We mainly use ColBERTv2~\citep{colbertv2} as the retriever, retrieving top-10 documents for each question. Alternately, BM25~\citep{robertson2009probabilistic}, Google  Search\footnote{\url{https://serper.dev/}} serve as retrievers in our analysis experiment. 
Following previous work~\cite{xu2023searchinthechain}, we use Wikipedia 2017 for HotpotQA, and a large-scale passage collection built on Wikipedia 2018 for other open-domain QA benchmarks.

Our instructional grounding distillation trains Mistral-7B with 50k synthetic examples.
We optimize the model using the deepspeed ZeRO  strategy~\citep{deepspeed} with the learning rate of $5e^{-5}$ and the weight decay coefficient of 0.01.
The training of our model can be done within 18 hours with 3 NVIDIA A100-PCIE-80GB GPUs.

\section{Experimental Results}

\subsection{Experimental Results}

\paragraph{Overall performance.} 
Table~\ref{tab:main} presents the experiment results. Our framework surpasses others on all four datasets and all metrics. Specifically, on the HotpotQA dataset, our \ours achieves Acc=47.27, F1=52.26, and Acc$\dagger$=55.73, considerably improving over the \textit{Generate w/ Retriever} baselines. It also significantly outperforms retrieval-then-read baselines like DSPy and SearChain, with a 4-6 point increase in accuracy metrics across all datasets. The similar improvement is observed in our human evaluation results (see Appendix \uppercase\expandafter{\romannumeral4} for more detials). 
These results indicate that our method effectively utilizes world knowledge and LLMs' deductive abilities to answer questions.

\paragraph{Results with the smaller model.}
We further evaluate our method by swapping the backbone LLMs with the open source model, \ie Mistral-7B, and repeating the experiment under the same conditions.
As shown in Table~\ref{tab:open-model}, we implement our methods in two ways with the Mistral-7B:
(1) directly prompting (vanilla);
(2) tuning with our proposed instructional grounding distillation (IGD).
We obverse that directly prompting Mistral-7B with our method yields better performance compared with baselines.
The instructional grounding distillation further improves overall performance significantly, \eg pushing the Acc to 42.08 in the HotpotQA dataset (9.84\% relative improvement) and 14.37 in the MusiqueQA dataset (26.4\% relative improvement).

\begin{table}[!t]
\centering
\begin{adjustbox}{width=\columnwidth,center}
\begin{tabular}{p{2cm} ccc c}
\toprule
\textbf{Method}
& \textbf{HQA}
& \textbf{MQA}
& \textbf{WQA}
& \textbf{Average  $\Delta\downarrow$} \\

\hline
\rowcolor{Gainsboro}\multicolumn{5}{l}{\textit{Retriever $\rightarrow$ BM25}} \\

\textbf{Ours} 
&  42.21 
& 18.32
&   40.32 
& -
\\

DSPy 
& 40.86
& 15.32
&  30.85
& 5.27$\downarrow$
\\

GRG  \textit{w/ dq}
& 41.31
& 15.62
& 38.84
& 2.36$\downarrow$
\\

RetGen 
& 39.12
& 8.41
& 35.83
&  6.50$\downarrow$
\\

SearChain
& 39.57
& 14.93
&  37.41
& 3.65$\downarrow$ 
\\

\hline
\rowcolor{Gainsboro}\multicolumn{5}{l}{\textit{Retriever $\rightarrow$ Google Search}} \\

\textbf{Ours}
& 48.95
& 21.54
& 46.87
& - 
\\

DSPy
& 46.86
& 20.71
& 39.92
& 3.29$\downarrow$ 
\\

GRG \textit{w/ dq}
& 42.57
& 18.41
& 43.21
& 4.39$\downarrow$ 
\\

RetGen 
& 42.82
& 14.27
& 44.31
&  5.32$\downarrow$ 
\\

SearChain 
& 44.35
& 19.76
& 44.39
&   2.95$\downarrow$
\\

\bottomrule
\end{tabular}
\end{adjustbox}
\caption{Accuracy (Acc) on three datasets using BM25 and Google Search as retrievers, respectively.
The \textit{w/ dq} is short for  \textit{without decomposition}.}
\label{tab:retriever}
\end{table}

\paragraph{Results with different retrievers.}
We further evaluate the performance of our framework by using different retrievers in various retrieval scenarios. As shown in Table~\ref{tab:retriever}, we replace ColBERTv2 with BM25 and Google Search, using ChatGPT as a backbone LLM in all instances. Our method demonstrates the best performance regardless of the retriever used, indicating its adaptability in both low recall (BM25) and high recall (Google Search) scenarios. This could be due to our answer deduction phase, which uses LLMs' parametric knowledge to supplement the retrieved knowledge. Moreover, our instructional knowledge grounding phase effectively incorporates the retrieved document by citing the most relevant evidence, mitigating the negative impact of noisy documents.

\subsection{Ablation Study}
We employ the following modifications and repeat the experiment in the same setting as Table~\ref{tab:main}.

\noindent \textbf{w/o deduction.}
We remove the \textit{answer deduction} phase mentioned in Section~\ref{sec:generation}, prompting the LLMs to directly generate an answer for a multi-hop question and revise it.
As illustrated in Table~\ref{tab:ablation}, we observe 6 and 10 absolute decreases in HQA and SQA datasets in terms of the Acc metric, respectively.
These results demonstrate the necessity of deducing the intricate knowledge in our answer deduction phase.

\noindent \textbf{w/o grounding.}
We replace the instructional knowledge grounding phase mentioned in Section~\ref{sec:grounding}) with directly generating an answer using retrieved documents
As shown in Table~\ref{tab:ablation}, the F1 and Acc metrics have a significant decline.
A potential reason is that the LLMs may hallucinate the plausible-looking statement in the retrieved documents. 
Our instructional knowledge grounding method further instructs the LLMs to find the most relevant evidence.

\noindent \textbf{w/o batch.}
We remove the \textit{batch grounding} strategy in Section~\ref{tab:main}.
As shown in Table~\ref{tab:ablation}, the F1 decreases from 52.26 to 47.27 and the Acc decreases from 47.27 to 45.03. 
These comparisons indicate that the LLMs struggle to generate correct answers when the reference document list is lengthy with irrelevant information.

\subsection{Case Study}

We conduct several case studies and find that \ours is more effective at generating high quality answers to a question. 
Details can be found in Appendix \uppercase\expandafter{\romannumeral2}.

\section{Analysis and Discussion}\label{sec:discussion}

\subsection{Result Consistency and Stability}
We further explore the consistency and stability of our framework.
Specifically, we repeat the experiment with the same setting as Table~\ref{tab:main} in the HotpotQA dataset.
The statistical significance of differences observed between the performance of two runs is tested using a two-tailed paired t-test.
We find no significant difference between the results of two randomly conducted experiments (significance level $\alpha$ = 0.05).
We further explore the \textit{fine-granularity stability} for our \textit{answer deduction} and \textit{instruction knowledge grounding} phases. Specifically, we compute the Rough score for the trajectory of our method in two repeated runs. The Rough-1, 2, and L are 81.33, 53.7, and 79.7, which shows the high lexicon similarity for the output, indicating the stability of our method.

\begin{figure}[t]
        \centering
\includegraphics[width=0.45\textwidth]{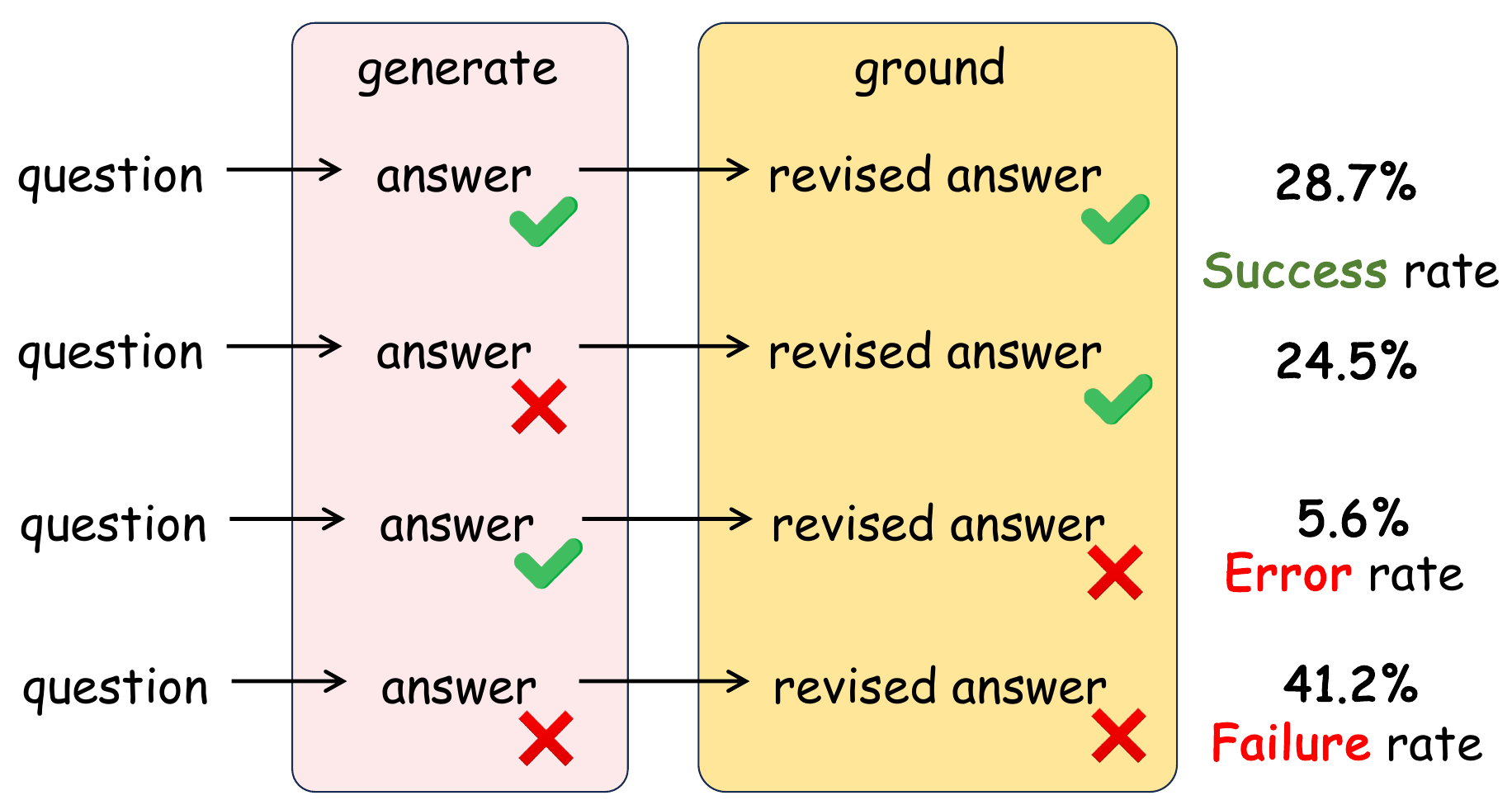}
        \caption{The fine-granularity \textit{correctness} analysis of our answer deduction and knowledge grounding phases.}
 \label{fig:correct}
\end{figure}

\subsection{Knowledge Incorporation}

Our method combines the knowledge in LLMs' parameters and external documents to answer a question. We explore the synergistic integration of these two distinct knowledge sources. Specifically, we calculate the following three metrics:
(1) \textit{Success rate}: the rate at which LLMs either directly generate a correct answer or accurately revise an incorrectly generated answer;
(2) \textit{Failure rate}: LLMs generate a wrong answer and fail to correct it;
(3) \textit{Error rate}: LLMs generate a correct answer but incorrectly revise it.

Our method addresses multi-hop questions step-by-step. As existing datasets lack the ground truth answer for immediate answering trajectory, we invite three annotators to evaluate 100 randomly sampled cases from the Hotpot QA dataset in Table~\ref{tab:main}.

As illustrated in Figure~\ref{fig:correct}, the overall \textit{success rate} is 53.2\%, with LLMs directly answering 28.7\% of questions correctly. For 24.5\% of the questions though, LLMs initially generate non-factual statements, and then use external documents for revision. These results underscore the importance of incorporating both knowledge sources.
During our grounding phase, LLMs may be misled by plausible-looking statements in the retrieved documents. Therefore, we further calculate the \textbf{error rate}, which assesses how often LLMs are incorrect after revisions. We find that the error rate is only 5.6\%, indicating that LLMs usually use the retrieved documents effectively.

\begin{figure}[!t]
        \centering
\includegraphics[width=0.45\textwidth]{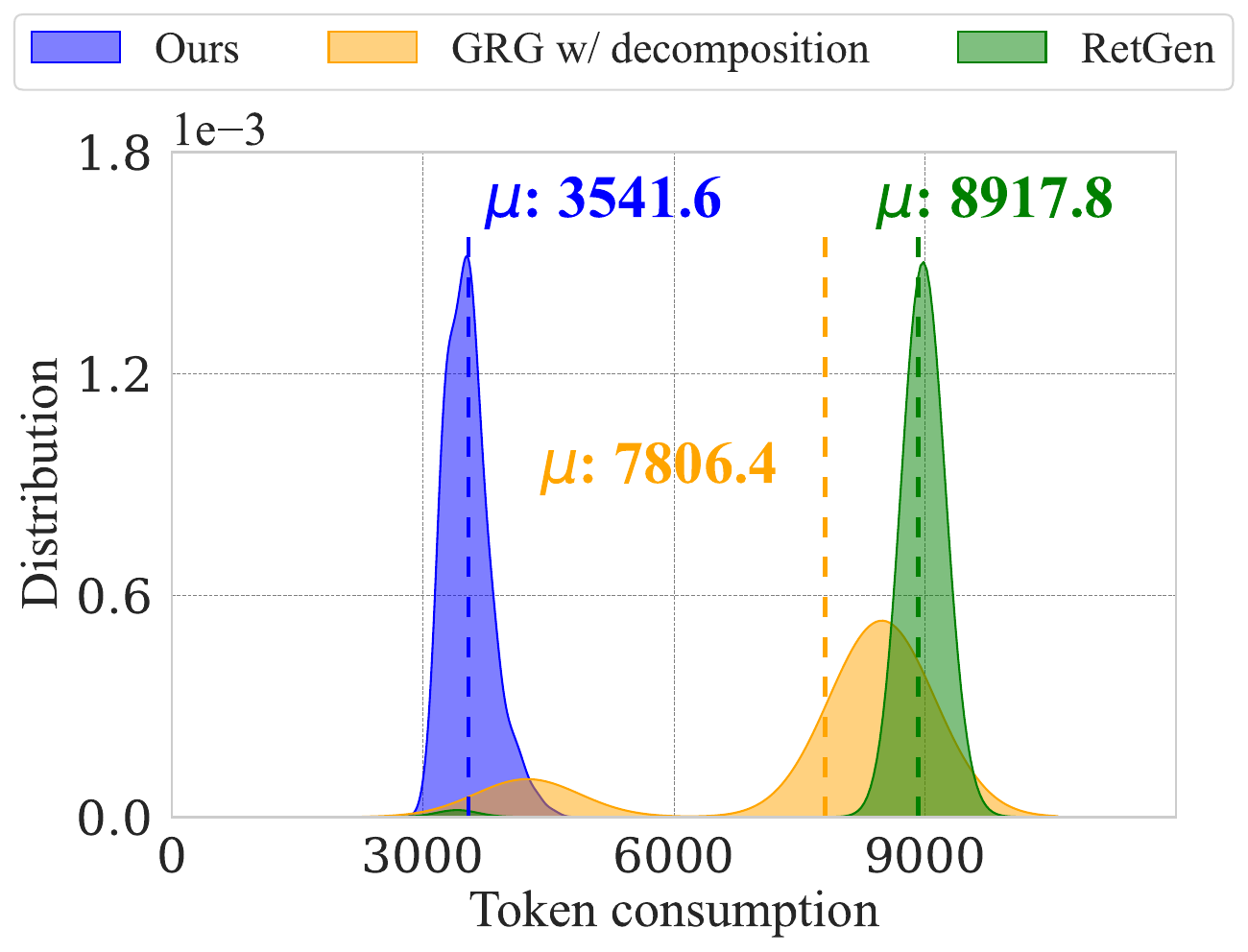}
        \caption{The efficiency analysis for different methods, where we count the number of consumed tokens and compute the average consumption $\mu$.}
 \label{fig:token}
\end{figure}

\subsection{Qualitative Analysis for Efficiency}
The intensive parameters of LLMs typically raise concern about inference cost. Thus, we compare \textbf{token consumption} with  \textit{GRG w/ decomposition} and \textit{RetGen}, using the HotpotQA dataset in Table~\ref{tab:main}.
We show the frequency histogram for the number of consumed tokens in different methods in Figure~\ref{fig:token}.
Though our framework achieves better performance, we observe that our method spends fewer tokens compared with strong baselines \textit{RetGen} and \textit{GRG w/ decomposition}.
The potential reason is that our framework benefits from the deduction capability of LLMs to decompose a multi-hop question into simpler sub-questions and generate an answer directly, leading to a shorter reasoning trajectory.

We also train Mistral-7B using different amounts of randomly sampled examples to investigate the impact of \textbf{data scale} on the effectiveness of our instructional grounding distillation (IGD). We notice a slight decrease in performance as the amount of data reduces. For instance, when training with 45k examples, our method achieves Acc=42.08; but with 20k examples, it achieves Acc=40.12. All results are averaged over three runs. We also observe that our IGD allows Mistral-7B to achieve performance comparable to, and sometimes even better than, strong baselines using ChatGPT as the backbone, such as ReAct.

\section{Conclusion}~\label{sec:conclusion}
We present a generate-then-ground (\ours) framework  for multi-hop question answering tasks, synergizing the parametric knowledge of LLMs and external documents to solve a multi-hop question.
Given a multi-hop question, our \ours enable LLMs to alternate two phases until the final answer is derived:
(1) formulate a simpler, single-hop question and directly generate the answer;
(2) ground the question-answer pair in retrieved documents, amending any wrong predictions in the answer.
To generalize our framework into smaller models, we also propose an instructional grounding distillation method.
Extensive experiments conducted on four datasets illustrate the superiority of our framework.

\section*{Limitations}\label{sec:lmitation}
Despite the promising results demonstrated in this paper, there are several limitations to our framework.
\begin{itemize}
    \item The first step of our framework involves generating an initial answer, which may be highly dependent on the task at hand. For different tasks, the model may struggle to generate a meaningful or useful initial answer, limiting its applicability across diverse problem domains.
    \item Our approach assumes that complex questions can be broken down into simpler ones. However, the task of decomposing complex questions is itself a challenging problem and has not been fully explored in our current framework.
    \item Our approach assumes that external documents can be used to correct initially non-factual statements generated by the model. However, if these sources do not contain the necessary information for correction, or if they contain misinformation, our framework's effectiveness could be compromised. 
\end{itemize}

\section*{Ethics Statement}

The research conducted in this paper centers around the development of a generate-then-ground framework for multi-hop question answering. Our framework enables Language Learning Models (LLMs) to alternately deduce answers and utilize established evidence to revise those answers across several iterations to solve multi-hop questions.

In the process of conducting this research, we have adhered to ethical standards to ensure the integrity and validity of our work. All the questions used in this study were obtained from existing benchmarks, thus ensuring a high level of transparency and reproducibility in our experimental procedure.

Furthermore, to support our retrieval system, we used an open source corpus, specifically, Wikipedia. This ensures that our research utilizes publicly accessible and freely available data, minimizing potential bias and promoting fairness.

We have made every effort to ensure that our research does not harm individuals or groups, nor does it involve any form of deception or potential misuse of information.

\section*{Acknowledgements}
This work was supported by the Natural Science Foundation of China (62102234, 62372275, 62272274, 62202271, T2293773, 62072279), the National Key R\&D Program of China with grant No.2022YFC3303004, the Natural Science Foundation of Shandong Province (ZR2021QF129).
All content represents the opinion of the authors, which is not necessarily shared or endorsed by their respective employers and/or sponsors.

\clearpage
\bibliography{custom}
\clearpage

\clearpage
\section*{Appendices}\label{app}

\subsection*{Appendix \uppercase\expandafter{\romannumeral1}. Evaluation Metrics Details}~\label{app:evaluate}
Following prior research~\citep{xu2023searchinthechain,Shao2023EnhancingRL,ren2023investigating}, we evaluate our method and baselines with three metrics:
(1) \textit{Accuracy} (Acc), which evaluates whether the ground truth answer is contained within the generated answer,
(2) \textit{F1 Score} (F1), which computes the lexical similarity between the generated answer and the ground truth answer. As the harmonic mean of precision and recall, it is calculated using term-level exact match other than ROUGE-L, with precision being the ratio of shared terms to predicted terms, and recall being the ratio of shared terms to actual terms,
and (3) \textit{Semantic Accuracy} (Acc$^\dagger$), which prompts the LLMs to evaluate the correctness of the generated answer taking the ground truth answer as reference.
In this work, we implement the Acc$^\dagger$ using gpt-3.5-turbo-instruct\footnote{\url{https://openai.com/}}. 
The prompt is as follows, where \code{\{question\}}, \code{\{model output\}}, and \code{\{answer\}} are placeholders.
The results are averaged over three runs.

\begin{tcolorbox}[colback=black!1!white,colframe=black!57!white,title=Prompt for Evaluating the Correctness of a Model Output]

In the following task, you are given a Question, a model Prediction for the Question, and a Ground-truth Answer to the Question. You should decide whether the model Prediction implies the Ground-truth Answer.\\
\\
Question\\
\{question\}\\
\\
Prediction\\
\{model output\}\\
\\
Ground-truth Answer\\
\{answer\}\\
\\
Does the Prediction imply the Ground-truth Answer? Output Yes or No:
\end{tcolorbox}\label{tab:prompt}

\subsection*{Appendix \uppercase\expandafter{\romannumeral2}. Case Study}\label{sec:app:case}

We conduct several case studies and find that our method is more effective at generating high-quality answers to a question. 
A concrete example is shown in Table~\ref{tab:case}. 
We find that our \ours can derive the correct answer successfully in two hops while the other baselines fail.
In the first hop, the \ours enables the LLMs to formulate a simpler, single-hop question and directly generate a correct answer, which intuitively demonstrates the world knowledge of LLMs.
We also observe that although a wrong prediction is generated in the second hop initially, our \ours can instruct LLMs to revise it with the assistance of an external document.
This phenomena further illustrates the necessity to incorporate both the parametric knowledge of LLMs and external documents to answer a complex, multi-hop question.

\subsection*{Appendix \uppercase\expandafter{\romannumeral3}. Training Dataset for Grounding Distillation}\label{sec:app:data}

\paragraph{Synthesize the dataset }
Our instructional grounding distillation collects the trajectory of LLMs, \ie ChatGPT, during the instructional knowledge grounding (Section~\ref{sec:grounding}).
To achieve this, we first sample 50k questions from the Natural Questions (NQ) dataset~\citep{kwiatkowski2019natural}.
Each question is paired with a corresponding ground-truth document and the noise documents.
Next, we supplement each question with an immediate answer and a detailed revision trajectory.
The immediate answer is generated directly by feeding the question into a smaller model, such as Mistral-7B.
The revision trajectory is generated by ChatGPT with the assistance of the ground truth document.

For practical consideration, we prompt the LLMs to encapsulate the cited evidence and revised answer with special tokens, \ie ``<ref> </ref>'' and  ``<revise> </revise>'' in the output.
We employ the following heuristic methods to filter the low-quality generated output:
\begin{itemize}
    \item The output contain no evidence, \eg ``<ref> Empty </ref>''.
    \item The output contain no revised answer, \eg no ``<revise>'' or ``</revise>'' can be found.
    \item The revised answer in the output is misaligned with the ground truth evidence.
\end{itemize}

We demonstrate diversity in the length of the input instruction, and output trajectory in Table~\ref{tab:dataset}.

\begin{table}
\centering
\begin{adjustbox}{width=\columnwidth,center}
\begin{tabular}{l cccc }
\toprule

& \textbf{Ours}
& \textbf{GRG}
& \textbf{DSPy}
& \textbf{SearChain}
\\

\toprule
\textbf{Correctness}
& 52.75
& 49.71
& 51.24
& 46.30

\\

\bottomrule
\end{tabular}
\end{adjustbox}
\caption{The result of our human evaluation. }
\label{tab:human}
\end{table}

\subsection*{Appendix \uppercase\expandafter{\romannumeral4}. Human Evaluation}\label{sec:app:human}

we conduct a human evaluation, with three educated individuals assessing the \textit{correctness} of 120 randomly sampled cases from four benchmarks on a three-scale rating.
Specifically,  the annotators are invited to manually evaluate the question-answering trajectory step-by-step using the ground truth document provided by existing datasets.
To guarantee annotation quality, we ask at least two annotators to evaluate the same questions repeatedly. 
If there is a discrepancy between the two annotators (\ie two annotators give a different correctness score), we ask a third annotator to recheck it. 
As the results are shown in Table~\ref{tab:human}, our method achieves the best performance with a 3.66 point improvement on average.
This result demonstrates the superiority of our method.
The overall Kappa value of our human evaluation is 0.73, illustrating agreement among the annotators.

\begin{table*}[htbp]
\small \centering
\setlength\tabcolsep{3.5pt}
\begin{tabular}{@{}p{15.5cm}@{}}
\toprule
\multirow{1}{*}{\textbf{Query}}
In what month is the annual documentary film festival, that is presented by the fortnightly published British journal of literary essays, held? (HotpotQA dataset)
\\

\hline
\rowcolor{Gainsboro} \multicolumn{1}{l}{\ours}
\\
\begin{tabular}[l]{@{}p{15.5cm}@{}}
Question 1: What is the name of the annual documentary film festival presented by the fortnightly published British journal of literary essays? \\
Answer 1:  The \textcolor{blue}{Fortnightly Review Documentary Film Festival}. \\
Grounding 1: The document demonstrate <ref> The annual documentary film festival presented by the fortnightly published British journal of literary essays is called the London International Documentary Festival (LIDF) </ref>. <revise>the London International \textcolor{blue}{Documentary Festival} (LIDF) </revise>. \\
Question 2: The annual documentary film festival presented by the fortnightly published British journal of literary essays is called the London International Documentary Festival (LIDF). In what month is LIDF held? \\
Answer 2: LIDF is held in the months of \textcolor{blue}{March and April*} every year. \\
Grounding 2: The document demonstrate <ref> The London International Documentary Festival (or LIDF) is an annual documentary film festival that takes place in the months of March and April every year </ref>. The revised answer is <revise> LIDF is held in the months of \textcolor{blue}{March and April} every year </revise>. \\
\#\#\#Finish[\textcolor{blue}{March and April}] 
\end{tabular} 
\\

\hline
\rowcolor{Gainsboro} \multicolumn{1}{l}{Chain of thought}
\\
\begin{tabular}[l]{@{}p{15.5cm}@{}}
Step 1: The annual documentary film festival is presented by \textcolor{blue}{the fortnightly} published British journal of literary essays. \\
Step 2: The entity presenting the festival is the British journal of literary essays.\\
Step 3: \textcolor{blue}{"Fortnightly"} means the journal is published every two weeks. \\
Step 4: Since the journal is published every two weeks, we need to find out which months fall around the time when the festival is typically held. If the journal is published fortnightly, it means there are approximately 26 issues per year. Given that the festival is annual, it should occur once within those 26 fortnightly periods.\\
Final Answer: This film festival, presented by the British journal of literary essays, is held in the month of \textcolor{red}{May}.
\end{tabular} 
\\

\hline
\rowcolor{Gainsboro} \multicolumn{1}{l}{GRG w/ decomposition}
\\
\begin{tabular}[l]{@{}p{15.5cm}@{}}
Question 1:  I need to get for Name of the British journal of literary essays\\
Document 1: [1] The \textcolor{blue}{London Review of Books (LRB)} is a British journal of literary essays.  It is published fortnightly... |||  [2]  Essays in London and Elsewhere is a book of literary criticism by Henry James published in 1893...\\
Question 2: When is the \textcolor{blue}{annual documentary film festival} held?
Document 2: The annual documentary film festival, also known as DOC NYC, is held every \textcolor{red}{November} in New York City. It was founded in 2010 and has since become one of the largest documentary film festivals in the United States. The festival showcases over 300 films and hosts panels, workshops. ||| The London International Documentary Festival (or LIDF) is an annual documentary film festival that takes place in the months of March and April every year.  The event...\\
\#\#\#Finish[\textcolor{red}{November}] \\
\end{tabular} 
\\

\hline
\rowcolor{Gainsboro} \multicolumn{1}{l}{SearChain}
\\
 \begin{tabular}[l]{@{}p{15.5cm}@{}}
Question 1: [Unsolved Query]: What is the name of the annual documentary film festival presented by the fortnightly published British journal of literary essays? [Unsolved Query]: What is the schedule or dates for the annual documentary film festival presented by the fortnightly published British journal of literary essays? [Unsolved Query]: When does the annual documentary film festival presented by the fortnightly published British journal of literary essays usually take place?
[Unsolved Query]: In what month is the annual documentary film festival usually held?\\
Document 1:  [1] Essays in London and Elsewhere is a book of literary criticism by Henry James published in 1893... \\
Answer 1: The annual documentary film festival presented by the \textcolor{blue}{fortnightly published British} journal of literary essays is called the London International Documentary Festival (LIDF).
Question 2: [Query]: When does the London International Documentary Festival usually take place? Answer 5: The London International Documentary Festival usually takes place in the months of \textcolor{red}{October} every year.
\\
\end{tabular} 
\\

\bottomrule
\end{tabular}
\caption{A concrete example on the HotpotQA dataset. We highlight the \textcolor{blue}{right} and \textcolor{red}{wrong} prediction used to answer the input question.
We find that our \ours can derive the correct answer successfully in two hops while the other baselines fail.
In the first hop, the \ours enables the LLMs to formulate a simpler, single-hop question and directly generate a correct answer.
We also observe that although a wrong prediction is generated in the second hop initially, our \ours can instruct LLMs to revise it with the assistance of an external document.}
\label{tab:case}
\end{table*}

\end{document}